\documentclass[9pt,conference]{IEEEtran}

\ifCLASSINFOpdf
\else
\fi
\usepackage{hyperref}
\usepackage{amsmath}
\usepackage{amssymb}
\usepackage{graphicx}
\usepackage{bm}
\usepackage{color}

\newcommand{\bb}[1]{\bm{\mathrm{#1}}}
\DeclareMathOperator*{\argmin}{arg\,min}

\graphicspath{ {./} }

\begin{document}
%
\title{Cloud Dictionary: \\ Coding and Modeling for Point Clouds}

\author{\IEEEauthorblockN{Or Litany$^*$}
\IEEEauthorblockA{orץlitany@gmail.com}
\and
\IEEEauthorblockN{Tal Remez$^*$}
\IEEEauthorblockA{talremez@gmail.com}\\
Tel-Aviv University\\
\tiny{$*$ Equal contributors}
\and
\IEEEauthorblockN{Alex Bronstein}
\IEEEauthorblockA{bron@eng.tau.ac.il}
}


%

\maketitle

\begin{abstract}
With the development of range sensors such as LIDAR and time-of-flight cameras, 3D point cloud scans have become ubiquitous in computer vision applications, the most prominent ones being gesture recognition and autonomous driving. Parsimony-based algorithms have shown great success on images and videos where data points are sampled on a regular Cartesian grid. We propose an adaptation of these techniques to irregularly sampled signals by using continuous dictionaries. We present an example application in the form of point cloud denoising.    
\end{abstract}

\IEEEpeerreviewmaketitle

\section{Introduction}
Recent advances in sensors introduce a wide family of data which do not lie on a regular grid. This can be either due to the lack of a temporal grid as in the event camera introduced in \cite{brandli2014240}, or a spatial one as is the case of point clouds produced, e.g., by LIDAR scanning the world in a varying angular velocity. 
These data serve as the input to a vast range of computer vision application such as 3D scene reconstruction \cite{newcombe2011kinectfusion}, or understanding \cite{Litany2016}. Sparse coding techniques have been shown to achieve state-of-the-art performance in many tasks for regularly sampled data such as images and videos. However, these methods do not apply directly to unstructured data like point clouds. 
It would therefore be beneficial to bridge the gap and allow the use of these techniques for irregularly sampled signals. To this end, we suggest a generalization of patch-based sparse representation and dictionary learning techniques for irregularly sampled data
, and show an application for point clouds denoising.

\section{Sparse coding for point clouds}
Patch-based processing of images and videos has been shown to be more computationally efficient  and to produce better results compared to global models.
Unfortunately, the definition of patches on a 3D shape is non-trivial. In \cite{digne2014self}, it was proposed to define a patch straightforwardly by finding all points inside a ball with Euclidean or, better, geodesic radius $r$, centered around a point. A similar approach was also used by \cite{elmoataz2008nonlocal,lozes2015pde}. A plane is subsequently fitted into this set of points using PCA, onto which the points are projected to yield a local system of coordinates $(u,v,w)$, with $w_i$ denoting the normal displacement of point $i$ from the plane. The main gap left in order to be able to use standard sparse coding algorithms is that the point locations $\bb{u}=(u,v)$ are not situated on a regular grid. The method in \cite{digne2014self} bridges this gap by a nearest-neighbors interpolation of the $w$ values on a predefined regular grid. However, this adds computational burden and introduces unnecessary sources of error. We choose, instead, to handle the data more naturally as sampled points of a latent continuous function. To this end, we model the points of each patch as
$\bb{y}_{\bb{G}} = w(\bb{G})+n(\bb{G})$, where the signal $w$ and the noise $n$ are now continuous signals sampled at locations $\bb{G}=\{\bb{u}_1,\dots,\bb{u}_l\}$. In this notation, denoising boils down to a pursuit problem of the form
\begin{eqnarray}
\argmin_{\bb{z}} \frac{1}{2}||\bb{y}_{\bb{G}}-\bb{D}(\bb{G})\bb{z}||_2^2 \mbox{ s.t. } \|\bb{z}\|_0 \leq L,
\label{eq:grid_less_persuite}
\end{eqnarray}
where $L$ is a parameter constraining the number of non-zero elements in the representation, and, with some abuse of notation, $\bb{D(G)z} = \mathrm{d}_1(\bb{G}) z_1 + \cdots + \mathrm{d}_M(\bb{G}) z_M$ denotes the synthesis of a continuous signal from a discrete combination of continuous dictionary atoms $\mathrm{d}_i(\bb{u})$. 
%
%
This formulation eliminates the necessity of a fixed grid for all signals. In other words, given a set of samples $\bb{y}_{\bb{G}}$ and a continuous dictionary $\bb{D}$, we can sample its continuous atoms at the locations $\bb{G}$, and proceed with a pursuit algorithm of our choice. Once the code $\bb{z}$ is obtained we can get an estimate of the continuous signal $\hat{w}(\bb{u}) = (d_1(\bb{u}), \cdots, d_M(\bb{u})) \bb{z}$.

\section{Dictionary learning}
So far we have assumed to be given a continuous dictionary. Such a dictionary can be either axiomatic (e.g., cosine or sine functions) or learned. Specifically, research on images showed a significant gain in performance for dictionaries that are learned from the data. In what follows, we propose a continuous dictionary learning algorithm.

Our generalized formulation of (\ref{eq:grid_less_persuite}) enables the adaptation of dictionary learning techniques, such as $k$-SVD, to grid-less data in the following way: 
We first fix a set of continuous functions $\bb{\phi}^{\mathrm{T}} = \{\mathit{\phi}_1,\mathit{\phi}_2,...,\mathit{\phi}_N\}$. Each atom in the dictionary $\bb{D}$ can be defined as a linear combination of these functions,
\begin{eqnarray}
\mathit{d}_m(\bb{u})=\bb{\phi}(\bb{u})^{\mathrm{T}} \bb{a}_m=\sum_{i=1}{}{\mathit{\phi}_i(\bb{u})\bb{A}_{im}}   ; & \bb{A} \in R^{N\times M}.
\end{eqnarray}
Let $\bb{G}_i$ be the grid of the $i$-th training patch, and let $\bb{y}_i = y_i(\bb{G}_i)$ be the corresponding patch values. When solving the pursuit problem (\ref{eq:grid_less_persuite}),
notice that since each data point is constructed from an arbitrary number of sample points $\bb{G}_i$, the data vectors $\bb{y}_i$ vary in size, yet are represented with the same number of coefficients in the dictionary $\bb{D}(\bb{G}_i)$. Following the $k$-SVD algorithm in \cite{Elad2006KSVD}, we initialize the dictionary using a random set of coefficients $\bb{A}$.
At each step we begin by finding a representation $\bb{z}_i$ for each of the training data vectors $\bb{y}_i$ by solving (\ref{eq:grid_less_persuite}).
During the dictionary-update stage, for each atom $m=1,\dots,M$

\begin{enumerate}
	
	\item Find the 
    examples that use the atom $\mathit{d}_m$.  $\Lambda_m = \{ i : \bb{z}_{im} \neq 0 \}$
	
	\item Minimize the residual, restricted to the subset $\Lambda_m$
    
    \item Repeat the above steps until the representation error of all examples is below some threshold
\end{enumerate}
Code is available at \url{github.com/orlitany/Cloud_dictionary}.

\section{Experiments}
\begin{figure}
	\centering
	\begin{tabular}{ c c c}		

		\includegraphics[width = 0.23\textwidth]{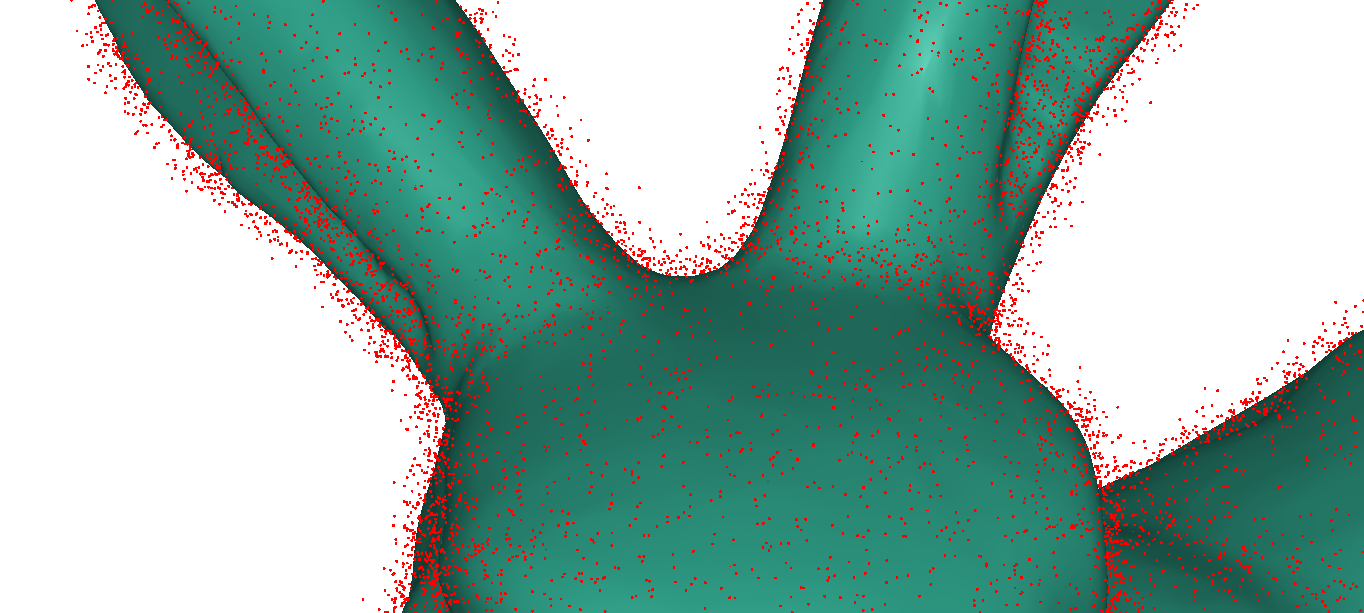} &		
		\includegraphics[width = 0.23\textwidth]{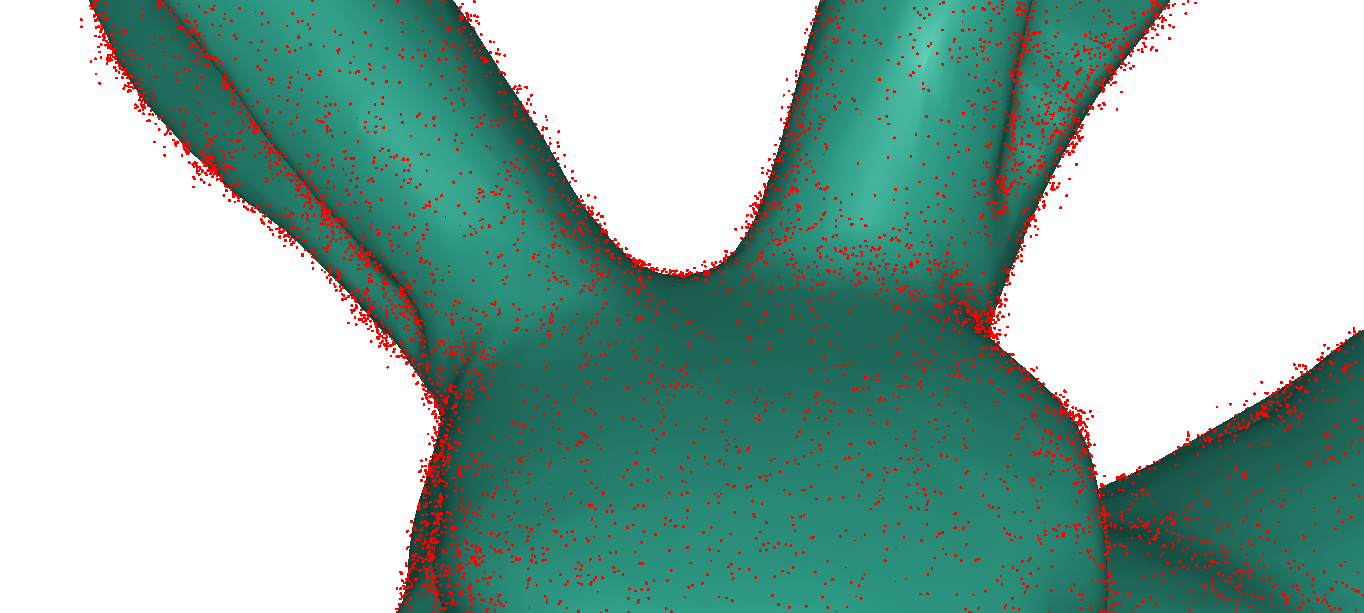} \\
		\includegraphics[width = 0.23\textwidth]{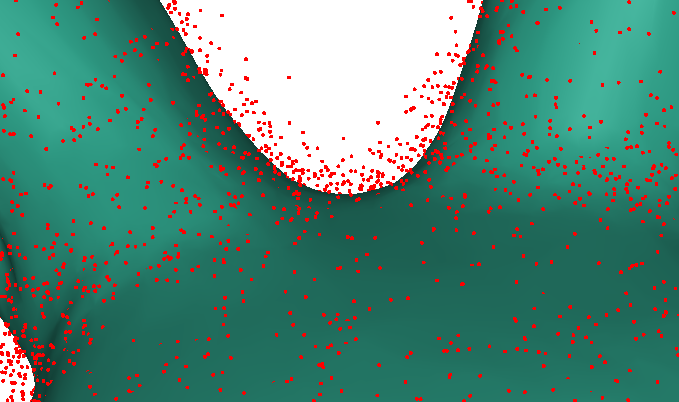} &		
		\includegraphics[width = 0.23\textwidth]{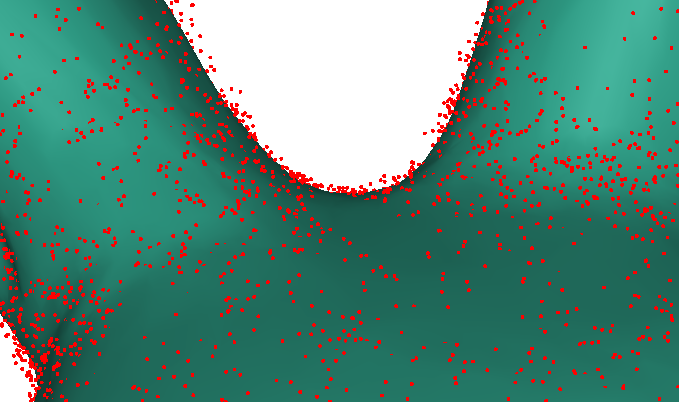} \\
		Noisy input & Our denoising \\

	\end{tabular}   \\
	\caption{ \textbf{Point could denoising using a continuous dictionary.} The images present a zoomed fraction of the bunny, where the ground truth mesh is presented in green and the point cloud in red. On the left is the noisy point cloud, and on the right is the denoised point cloud using our denoising algorithm. These results clearly show that our denoising algorithm returns a cleaner point cloud that is closer to the ground truth mesh.}
	\label{fig:pcl_denoise}
\end{figure}

\begin{figure}\label{fig:dict_learning}
\includegraphics[width = 0.5\textwidth]{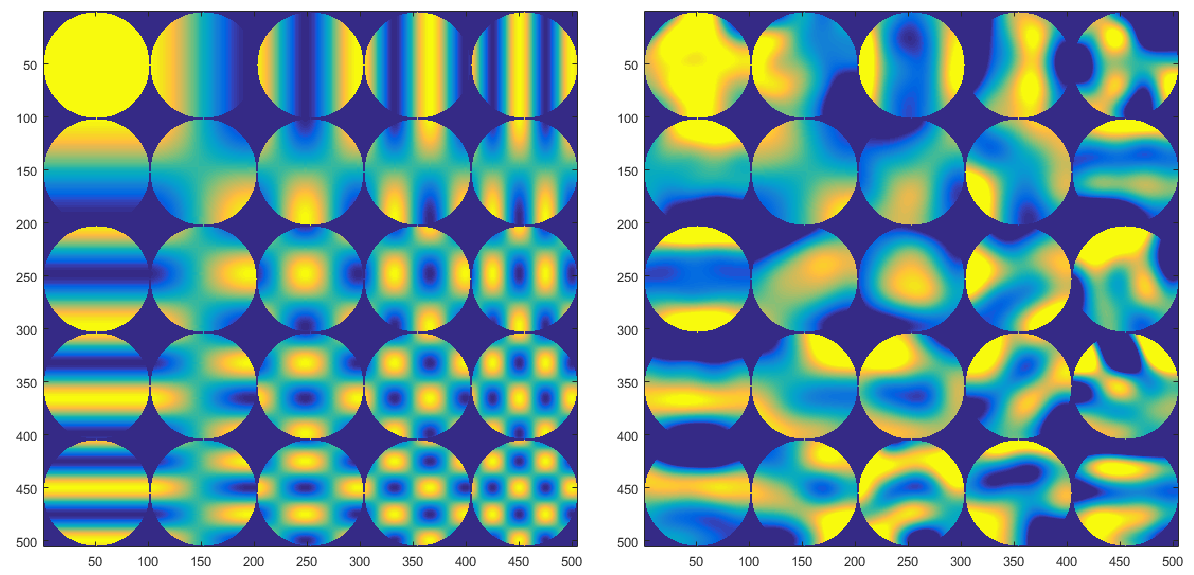}
\caption{\textbf{Continuous dictionary learning.} On the left is the original continuous dictionary atoms before the dictionary learning process, constructed using a Cartesian product of cosines. On the right is the learned continuous dictionary. It can be seen that the learned dictionary is smoother and introduces new structures at different rotation angles. The error per dictionary learning iteration is shown in Figure \ref{fig:dict_learning_error}.}
\end{figure}

\begin{figure}\label{fig:dict_learning_error}
\includegraphics[width = 0.5\textwidth]{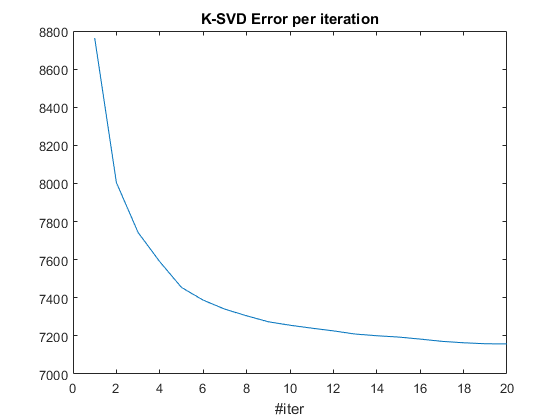}
\caption{\textbf{Continues $k-$SVD learning error Vs iteration.} Presented is the error of synthesizing the Bunny using the learned continuous dictionary after each $k-$SVD iteration.}
\end{figure}

We demonstrate the proposed method on the canonical example of denoising. Given a noisy point cloud (Figure \ref{fig:pcl_denoise}), we constructed a continuous dictionary composed of cosine functions with varying spatial frequencies and applied the proposed dictionary learning technique. Figure \ref{fig:dict_learning} shows the original and the learned dictionaries. The learning process error rate is presented in Figure \ref{fig:dict_learning_error}. We obtain the denoised point cloud (Figure \ref{fig:pcl_denoise}) by a convex relaxation of equation \ref{eq:grid_less_persuite}, and by averaging reconstructions from overlapping patches.

\begin{figure}
	\centering
    \begin{tabular}{ c}		
		\includegraphics[width = 0.3\textwidth]{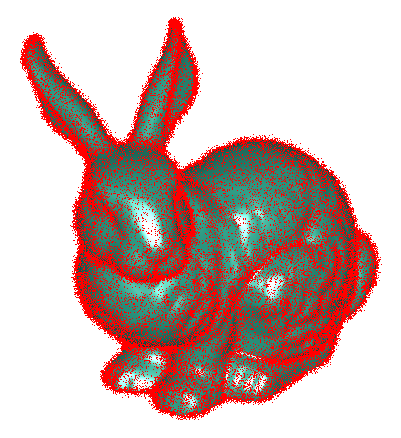} \\
	\end{tabular}   \\	
 	\caption{\textbf{Noisy point cloud of Stanford's bunny \cite{turk2005stanford}}. The original bunny mesh is presented in green and the noisy points cloud contaminated with Gaussian noise is presented in red.}
\label{fig:pcl_noisy}
\end{figure}

\bibliographystyle{IEEEtran}
\bibliography{pc_references}





\end{document}